%% file: main.tex
\documentclass[10pt,twocolumn,letterpaper]{article}

\usepackage[pagenumbers]{cvpr} 

\usepackage{amsmath}  
\usepackage{amsthm}  
\usepackage{graphicx}  
\usepackage{tcolorbox} 
\usepackage{multirow}
\usepackage{colortbl}
\usepackage{bbding}
\usepackage{wrapfig}
\usepackage{array}  

\usepackage{amssymb,booktabs}
\usepackage{tabularx}
\usepackage{bbding}
\usepackage{pifont}
\usepackage{makecell}
\usepackage{ulem}

\input{preamble}
\definecolor{cvprblue}{rgb}{0.21,0.49,0.74}
\usepackage[pagebackref,breaklinks,colorlinks,allcolors=cvprblue]{hyperref}

\def\paperID{*****} 
\def\confName{CVPR}
\def\confYear{2026}

\title{RemoteAgent: Bridging Vague Human Intents and Earth Observation with RL-based Agentic MLLMs}

\author{
\textbf{Liang Yao}\textsuperscript{1,*},
\textbf{Shengxiang Xu}\textsuperscript{2,*},
\textbf{Fan Liu}\textsuperscript{1,†},
\textbf{Chuanyi Zhang}\textsuperscript{1},
\textbf{Bishun Yao}\textsuperscript{1}\\
\textbf{Rui Min}\textsuperscript{1},
\textbf{Yongjun Li}\textsuperscript{1},
\textbf{Chaoqian Ouyang}\textsuperscript{3},
\textbf{Shimin Di}\textsuperscript{2},
\textbf{Min-Ling Zhang}\textsuperscript{2}
\\
\textsuperscript{1}Hohai University \quad
\textsuperscript{2}Southeast University \quad
\textsuperscript{3}Sun Yat-sen University \\
\\
{\tt\small \textsuperscript{*}Equal Contribution \quad \textsuperscript{†}Corresponding Author}
\\
{\tt\small Email: fanliu@hhu.edu.cn}
\\
{\tt\small GitHub Repo: \url{https://github.com/1e12Leon/RemoteAgent}}
}

\begin{document}
\maketitle
\input{sec/0_abstract}    
\input{sec/1_intro}

\input{sec/3_VagueEO}
\input{sec/4_RemoteAgent}

\input{sec/5_Experiments}

\input{sec/2_related_work}

\input{sec/6_Conclusion}
{
    \small
    \bibliographystyle{ieeenat_fullname}
    \bibliography{main}
}


\end{document}

%% file: sec/0_abstract.tex
\begin{abstract}
Earth Observation (EO) systems are essentially designed to support domain experts who often express their requirements through vague natural language rather than precise, machine-friendly instructions. Depending on the specific application scenario, these vague queries can demand vastly different levels of visual precision. Consequently, a practical EO AI system must bridge the gap between ambiguous human queries and the appropriate multi-granularity visual analysis tasks, ranging from holistic image interpretation to fine-grained pixel-wise predictions.
While Multi-modal Large Language Models (MLLMs) demonstrate strong semantic understanding, their text-based output format is inherently ill-suited for dense, precision-critical spatial predictions. Existing agentic frameworks address this limitation by delegating tasks to external tools, but indiscriminate tool invocation is computationally inefficient and underutilizes the MLLM's native capabilities.
To this end, we propose RemoteAgent, an agentic framework that strategically respects the intrinsic capability boundaries of MLLMs. To empower this framework to understand real user intents, we construct VagueEO, a human-centric instruction dataset pairing EO tasks with simulated vague natural-language queries. By leveraging VagueEO for reinforcement fine-tuning, we align an MLLM into a robust cognitive core that directly resolves image- and sparse region-level tasks. Consequently, RemoteAgent processes suitable tasks internally while intelligently orchestrating specialized tools via the Model Context Protocol exclusively for dense predictions. Extensive experiments demonstrate that RemoteAgent achieves robust intent recognition capabilities while delivering highly competitive performance across diverse EO tasks. 
\end{abstract}

%% file: sec/1_intro.tex
\section{Introduction}
\label{sec:intro}

\begin{figure}[t]
  \centering
  \includegraphics[width=\linewidth]{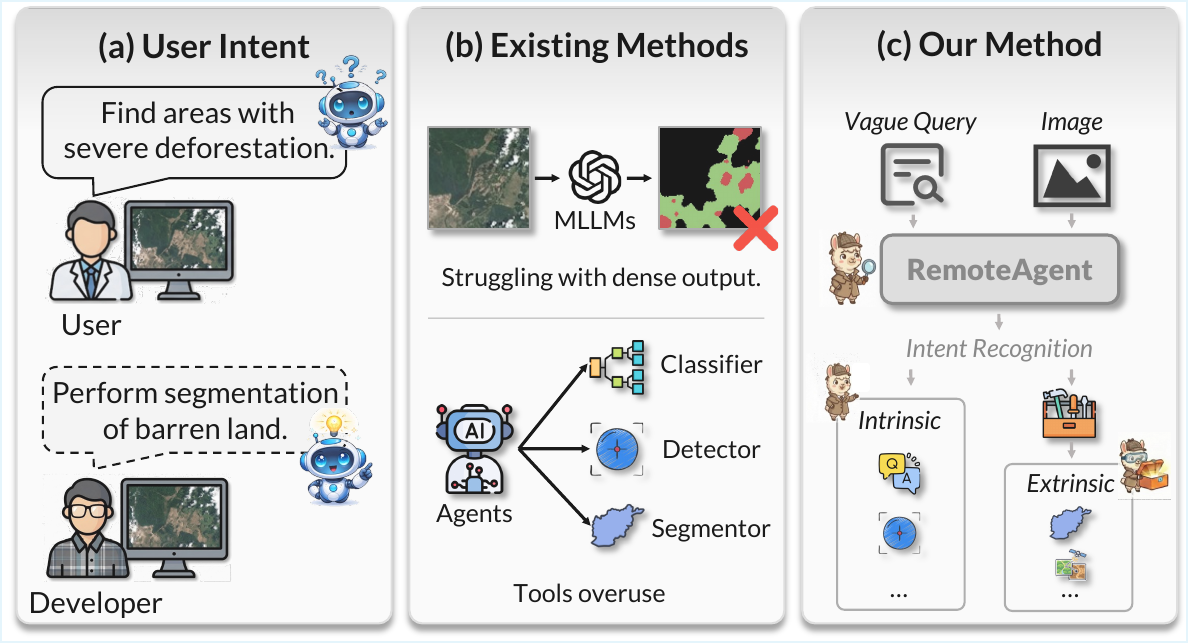}
  \caption{ 
(a) The usability gap between vague user intents and rigid system requirements. 
(b) Existing MLLMs struggle with dense output tasks, whereas tool-augmented agents suffer from indiscriminate tool overuse. 
(c) RemoteAgent bridges this gap by internally resolving macroscopic queries while orchestrating specialized tools strictly for dense predictions.
}
  \label{fig1}
\end{figure}

We are interested in constructing Earth Observation (EO) systems~\cite{zhang2024vision,weng2025vision,zhou2024towards,zou2025remotetrimmer,li2024unleashing} that achieve both rigorous precision and high practical utility. The true practical value of an EO system heavily relies on its accessibility to its primary end-users, domain experts such as earth scientists, urban planners, and policymakers. However, a critical usability gap hinders current deployments: these users typically lack the computer science background required to formulate machine-friendly instructions, such as strictly defined class taxonomies or explicit coordinate formats. Instead, they naturally express their analytical needs through vague, free-form language queries. For instance, as shown in Fig.~\ref{fig1}, a policymaker is more likely to simply ask a system to "\textit{find areas with severe deforestation}", rather than rigidly instructing it to "\textit{perform semantic segmentation of barren land}". Therefore, a highly practical EO agent must act as an intelligent bridge, capable of reliably grounding these ambiguous human intents into actionable visual operations. Crucially, to satisfy the requirement of rigorous precision, the tasks derived from such open-ended queries must dynamically span a wide spectrum of granularity, ranging from holistic image-level understanding to fine-grained, pixel-wise dense predictions~\cite{yao2025remotesam,liu2024rsunivlm,li2026rsvg}. Consequently, an ideal EO system must seamlessly integrate robust intent recognition with multi-granularity task execution ability.

Given the dual requirement to interpret vague, free-form queries and unify diverse EO applications within a single paradigm, Multi-modal Large Language Models (MLLMs) have naturally emerged as promising candidates~\cite{kuckreja2024geochat,yao2025falcon,hu2025rsgpt,muhtar2024lhrs,yao2025uemm}. However, relying on a monolithic MLLM to handle the entire spectrum of EO tasks exposes two major bottlenecks. First, their auto-regressive, text-based architecture is fundamentally unsuited for dense, precision-critical spatial outputs. Second, to adapt these general-purpose models to specialized remote sensing domains, existing approaches often rely on extensive Supervised Fine-Tuning (SFT)~\cite{zheng2025learning,xu2026parameter}. Unfortunately, this heavy reliance on SFT inevitably triggers catastrophic forgetting, eroding the model's intrinsic open-ended reasoning capabilities~\cite{yao2026remotereasoner}. Ironically, this degradation destroys the very semantic flexibility required to decipher the ambiguous human intents we initially aimed to support.

To bypass the structural limitations of MLLMs in dense spatial predictions, recent works~\cite{shabbir2026openearthagent,feng2025earth,chen2025cangling,bhattaram2025geoflow} increasingly adopt agentic frameworks. By delegating execution to specialized external tools, these systems relieve the MLLM from directly generating dense outputs. However, this tool-augmented paradigm often falls into the opposite extreme: an indiscriminate reliance on external tools for almost all tasks. 
Relying on external tools for all queries not only introduces unnecessary computational inefficiency but also fails to leverage the native proficiency of MLLMs in holistic image interpretation.
Furthermore, without specialized alignment for human-centric interactions, existing agents still struggle to robustly map vague, free-form user intents to the correct sequence of operations.
Therefore, a more elegant routing strategy is required~\cite{xu2025robustflow, wang2026learning}: one that delegates tasks to specialized tools only when strictly necessary, while maximizing the MLLM's intrinsic strengths.

Motivated by these observations, we propose RemoteAgent, an agentic framework designed to bridge the usability gap in remote sensing by strategically respecting the intrinsic capability boundaries of MLLMs. To empower this framework to comprehend authentic, free-form human intents, we construct VagueEO, a human-centric instruction dataset. Unlike traditional datasets~\cite{Zhou2026geochef}, VagueEO pairs standard computer vision-oriented EO tasks with simulated vague, natural-language queries that accurately reflect the needs of non-expert users. Rather than forcing the model into a monolithic role via standard Supervised Fine-Tuning (SFT), we utilize VagueEO for reinforcement fine-tuning. This paradigm adapts the MLLM exclusively to image- and sparse region-level tasks. This RL-based alignment endows the model with robust reasoning capabilities while avoiding the generalizability degradation typical of SFT, thereby preserving the MLLM as a smart cognitive core. Therefore, RemoteAgent executes a highly efficient task routing strategy: it directly resolves suitable macroscopic tasks internally, while intelligently orchestrating specialized external tools via the Model Context Protocol (MCP)~\cite{hou2025model,ouyang2025code2mcp,di2026toolrosetta} exclusively for dense, precision predictions. 
By disentangling intent understanding and sparse tasks from dense task execution, we establish a flexible and precise EO system tailored for real-world utility.

To comprehensively validate the efficacy of RemoteAgent, we evaluate it in three distinct dimensions: (1) Intent recognition, which measures the accuracy of grounding vague, free-form user queries into the correct operational pipelines. (2) Intrinsic capability, which assesses the RemoteAgent's native ability to directly resolve image-level and sparse region-level tasks. (3) Extrinsic execution, which evaluates its proficiency and accuracy in orchestrating specialized tools for dense predictions. 
Experimental results confirm that RemoteAgent accurately maps free-form user intents to correct pipelines. For intrinsic tasks, it delivers competitive performance with significantly less data than MLLMs. Finally, for extrinsic tasks, our routing mechanism substantially outperforms MLLM baselines, yielding spatial precision comparable to specialized models.
Our contributions are summarized as follows:
\begin{itemize}
    \item We address the disconnect between rigid EO benchmarks and free-form human intents by introducing VagueEO, a dataset to train and evaluate MLLMs on vague queries.
    \item We propose RemoteAgent, an agentic system that uses RL-alignment to resolve intrinsic tasks while routing dense predictions via specialized tools.
    \item Holistic experiments demonstrate that RemoteAgent achieves exceptional data efficiency on intrinsic MLLM tasks and expert-level precision on extrinsic tool invocations.
\end{itemize}

%% file: sec/3_VagueEO.tex
\section{VagueEO}
\begin{figure*}[t]
  \centering
  \includegraphics[width=\linewidth]{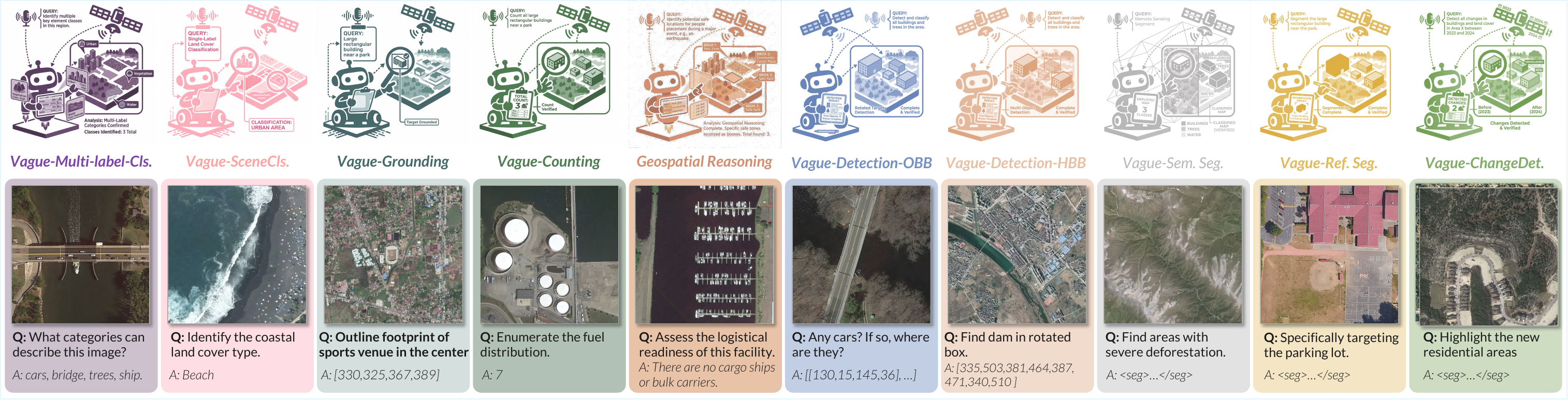}
  \caption{VagueEO Benchmark Overview. We construct ten diverse Earth Observation tasks that pair vague, human-centric queries with standardized structural annotations.}
  \label{data}
\end{figure*}

While recent remote sensing datasets have made remarkable strides in multi-modal alignment, they predominantly feature explicit, well-structured instructions. This paradigm inadvertently overlooks the inherent ambiguity and free-form nature of real-world queries from non-expert Earth Observation users. To bridge the gap between these machine-centric setups and real-world usability, we curate VagueEO, a dataset specifically designed to capture the linguistic ambiguity of non-expert queries, as shown in Fig.~\ref{data}.

We employ a scalable LLM-driven synthesis pipeline, which prompts LLMs to generate a diverse set of vague query templates that reflect real-world user intents. These simulated queries are then directly paired with high-quality structural annotations from standard Earth Observation benchmarks. Consequently, VagueEO features two key characteristics:

\begin{itemize}
    \item Free-form Natural Language: Instead of strictly formatted commands, the queries use everyday, ambiguous expressions (e.g., "can you point out any planes here?"). This explicitly forces the model to learn intent deduction rather than simple keyword matching.
    \item Multi-Granularity Annotations: Each vague query is mapped to precise visual ground truths in a deterministic manner. The annotations cover multiple spatial scales, ranging from image-level labels to bounding boxes and pixel-wise masks, providing the supervision needed for both semantic understanding and spatial reasoning.
\end{itemize}

We partition VagueEO into distinct training and testing sets. This split is specifically designed to train the MLLM's intent recognition on sparse tasks, while evaluating the framework's routing capability on unseen, dense spatial tasks.

Training Set (Intrinsic Tasks): Since general-purpose MLLMs have been widely proven to inherently excel at macroscopic and sparse understanding, we exclusively construct our training corpus around these intrinsic tasks. It consists of 5 tasks: Scene Classification, Multi-label Classification, Visual Grounding, Object Counting, and Geospatial Region Reasoning. We generate exactly 1,000 vague query-annotation pairs for each category. This set is used exclusively for the reinforcement fine-tuning of the MLLMs.

Testing Set (Intrinsic \& Extrinsic Tasks): The testing set evaluates the full system across 10 mainstream Earth Observation tasks. In addition to the 5 training tasks, it introduces 5 completely unseen tasks, predominantly featuring dense spatial predictions (e.g., Object Detection, Semantic Segmentation, Referring Expression Segmentation, and Change Detection). We construct 100 query-annotation pairs for all 10 tasks.

We hope VagueEO can provide the remote sensing community with a definitive benchmark to evaluate capability-aware routing.

%% file: sec/4_RemoteAgent.tex
\section{RemoteAgent}
\label{sec:method}

\begin{figure*}[t]
  \centering
  \includegraphics[width=\linewidth]{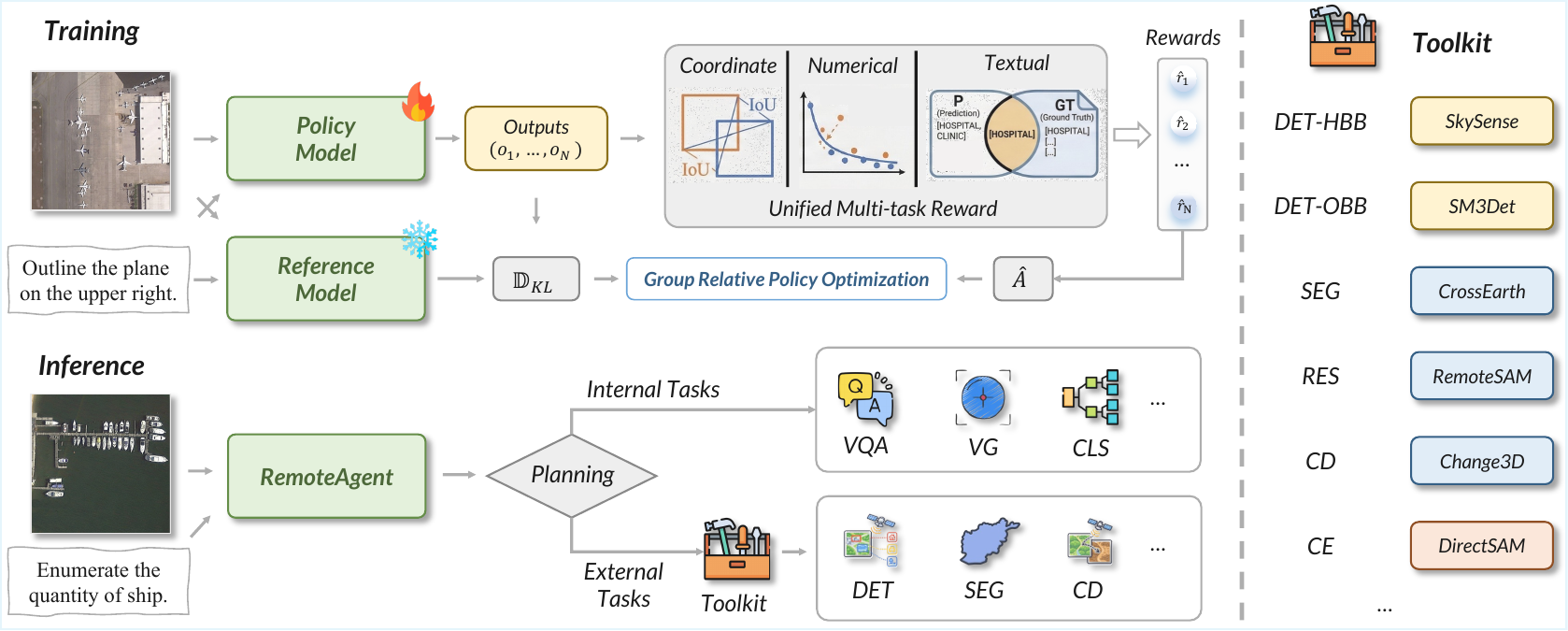}
  \caption{Overview of RemoteAgent. During training, the model is aligned via GRPO, guided by a unified multi-task reward that evaluates coordinate, numerical, and textual outputs. During inference, the agent dynamically plans and routes queries, directly resolving macroscopic tasks internally while delegating dense predictions to a specialized external toolkit. Task abbreviations: Visual Question Answering (VQA), Visual Grounding (VG), Classification (CLS), Detection (DET), Segmentation (SEG), Referring Expression Segmentation (RES), Change Detection (CD), and Contour Extraction (CE).}
  \label{overview}
\end{figure*}

We propose RemoteAgent in Fig.~\ref{overview}, which bridges vague user queries and precise EO tasks via an agentic framework. We detail the task formulation, training, and tool-augmentation in the following subsections.

\subsection{Formulation}

Given a remote sensing image $I$ and a task query $Q$, our goal is to learn a unified policy $\pi_\theta$ that generates an optimal response $O$. We categorize the task space $\mathcal{T}$ into two subsets based on the intrinsic suitability of MLLMs:
\begin{itemize}
    \item Intrinsic Tasks ($\mathcal{T}_{in}$): Semantic understanding and sparse reasoning tasks (e.g., Classification, visual grounding) where MLLMs excel.
    \item Extrinsic Tasks ($\mathcal{T}_{ex}$): Dense prediction tasks (e.g., segmentation, object detection) requiring pixel-level precision, handled by an external tool library $\mathcal{E}$.
\end{itemize}

The agent's output $O$ is formalized as a hybrid action space:
\begin{equation}
    O = \begin{cases} 
        R_{ans}, & \text{if } (I, Q) \in \mathcal{T}_{in} \\
        T_{call}(e_k, p), & \text{if } (I, Q) \in \mathcal{T}_{ex} 
        \end{cases}
    ,
\label{eq:actionspace}
\end{equation}
where $R_{ans}$ denotes the direct textual response, and $T_{call}(e_k, p)$ represents invoking a tool $e_k \in \mathcal{E}$ with parameters $p$ via the Model Context Protocol (MCP).

Instead of maximizing likelihood via SFT, we optimize $\pi_\theta$ using Group Relative Policy Optimization (GRPO) to maximize the expected reward $\mathbb{E}[r(O)]$, ensuring the model learns to autonomously distinguish between solving $\mathcal{T}_{in}$ internally and routing $\mathcal{T}_{ex}$ to tools while preserving general reasoning capabilities.

\subsection{RemoteAgent Training}


RemoteAgent builds on Qwen2.5-VL-7B-Instruct~\cite{bai2025qwen2} and is optimized as a multimodal policy $\pi_\theta$ over 5 intrinsic structured sparse reasoning tasks, including scene classification, multi-label classification, visual grounding, object counting, and region reasoning. For such intrinsic tasks, RemoteAgent directly generates a structured answer $R_{\mathrm{ans}}$ without invoking external dense prediction tools.

\subsubsection{GRPO-based Optimization}

To optimize $\pi_\theta$ for structured sparse visual reasoning, RemoteAgent adopts Group Relative Policy Optimization (GRPO)\cite{shao2024deepseekmath} instead of Supervised Fine-Tuning (SFT). Unlike SFT, which maximizes token-level likelihood and encourages imitation of reference phrasing\cite{wu2025generalization}, GRPO directly rewards the functional correctness of structured outputs and is therefore better aligned with the target objective. Combined with KL regularization, this formulation further constrains policy drift and helps retain the base model’s general capabilities during optimization~\cite{yao2026remotereasoner}. Crucially, this preserves its zero-shot ability to interpret system prompts and route dense tasks to external tools.

For each input pair $(I,Q)$, we sample $N$ outputs $\{o_i\}_{i=1}^{N}\sim \pi_{\theta_{\mathrm{old}}}(\cdot\mid I,Q)$ and assign each a scalar reward $r_i = R(I,Q,o_i)$. 
Rewards are standardized within each group to obtain normalized advantages
\begin{equation}
A_i = \frac{r_i - \mu_r}{\sigma_r},
\end{equation}
where $\mu_r$ and $\sigma_r$ denote the empirical mean and standard deviation of $\{r_j\}_{j=1}^{N}$, respectively.

Since rewards are defined at the sequence level whereas $\pi_\theta$ is autoregressive, the group-normalized advantage is broadcast to all generated tokens. Specifically, let $o_i = (o_{i,1}, \ldots, o_{i,T_i})$ denote the $i$-th generated sequence, and define the token-level context at position $t$ as $s_{i,t} = (I, Q, o_{i,<t})$. We then assign $\hat A_{i,t} = A_i$ for all generated tokens and optimize the policy using the clipped GRPO objective with KL regularization:

\begin{equation}
\mathcal{J}_{\mathrm{GRPO}}(\theta)
=
\mathbb{E}\!\left[
\frac{1}{N}\sum_{i=1}^{N}\frac{1}{T_i}\sum_{t=1}^{T_i}
\Big(
\mathcal{L}^{\mathrm{clip}}_{i,t}
-\beta\,\mathrm{KL}_{i,t}
\Big)
\right].
\end{equation}

Here, the clipped surrogate objective $\mathcal{L}^{\mathrm{clip}}_{i,t}$ is given by

\begin{equation}
\mathcal{L}^{\mathrm{clip}}_{i,t}=\min\!\big(\rho_{i,t}\hat A_{i,t},\ \mathrm{clip}(\rho_{i,t},1-\epsilon,1+\epsilon)\hat A_{i,t}\big),
\end{equation}
where $\rho_{i,t} = \frac{\pi_\theta(o_{i,t}\mid s_{i,t})}{\pi_{\theta_{\mathrm{old}}}(o_{i,t}\mid s_{i,t})}$ represents the probability ratio between the active policy and the previous behavior policy $\pi_{\theta_{\mathrm{old}}}$. The token-level penalty $\mathrm{KL}_{i,t} = D_{\mathrm{KL}}\!\big(\pi_\theta(\cdot\mid s_{i,t})\ \Vert\ \pi_{\mathrm{ref}}(\cdot\mid s_{i,t})\big)$ explicitly bounds the deviation from the frozen base model $\pi_{\mathrm{ref}}$.

\subsubsection{Unified Multimodal Reward}

We employ a unified multimodal reward that maps heterogeneous structured outputs into scalar rewards for GRPO. The evaluator operates solely on the content of the \texttt{<answer>} field and infers the scoring branch from the format of the reference answer, without relying on task labels. Given a prediction--ground-truth pair $(a_{\mathrm{pred}}, a_{\mathrm{gt}})$, where $a_{\mathrm{pred}}$ is extracted from the \texttt{<answer>} span of the model output and $a_{\mathrm{gt}}$ is obtained from the annotated solution, the reward is dispatched to one of three branches:
\begin{equation}
\resizebox{0.9\linewidth}{!}{%
$\displaystyle
R(a_{\mathrm{pred}},a_{\mathrm{gt}})
=
\begin{cases}
R_{\mathrm{coord}}(a_{\mathrm{pred}},a_{\mathrm{gt}}), & \text{coordinate tuples},\\
R_{\mathrm{num}}(a_{\mathrm{pred}},a_{\mathrm{gt}}),   & \text{scalar values},\\
R_{\mathrm{text}}(a_{\mathrm{pred}},a_{\mathrm{gt}}),  & \text{label strings}.
\end{cases}
$%
}
\end{equation}
Invalid or missing answer spans receive zero reward.


For \textbf{coordinate-valued answers}, as used in visual grounding and region reasoning, the predicted and reference answers are parsed into sets of axis-aligned bounding boxes $P$ and $G$. To ensure permutation invariance, we perform Hungarian matching on the pairwise IoU matrix and define

\begin{equation}
R_{\mathrm{coord}}(P,G)
=
\frac{1}{|G|}
\sum_{(g,p)\in \mathrm{match}(G,P)}
\mathrm{IoU}(g,p),
\end{equation}
which jointly accounts for localization quality and coverage.

For \textbf{numerical answers} in object counting, let $g$ denote the ground-truth value and $p$ the parsed prediction. We use a relative-error-based reward with hard rejection of large errors:


\begin{equation}
\resizebox{0.9\linewidth}{!}{%
$\displaystyle
R_{\mathrm{num}}(P,G)
=
\begin{cases}
1, & p = g,\\[3pt]
0, & (g, p \neq 0)\ \vee\ \left(\dfrac{|p-g|}{|g|} > 0.5\right),\\[6pt]
\mathrm{e}^{-3\,\dfrac{|p-g|}{|g|}}, & \text{otherwise}.
\end{cases}
$%
}
\end{equation}

For \textbf{textual answers} in classification, $a_{\mathrm{pred}}$ and $a_{\mathrm{gt}}$ are canonicalized into label sets $P$ and $G$. We define
\begin{equation}
R_{\mathrm{text}}(P,G)
=
\begin{cases}
0, & G \cap P = \varnothing,\\[3pt]
1, & G \subseteq P,\\[3pt]
\dfrac{|G \cap P|}{|G|}, & \text{otherwise},
\end{cases}
\end{equation}
which behaves as a coverage-based score for single-label cases and as recall in the multi-label setting.


All scoring branches map heterogeneous structured outputs into $[0,1]$, providing a unified scalar reward interface for GRPO. Because reward computation is dispatched according to answer format rather than task labels, the same evaluator can supervise scene classification, region reasoning, visual grounding, and object counting without introducing task-specific losses. By contrast, dense pixel-level predictions are handled by external expert tools.

\subsection{Tool-Augmented Inference}
Once the policy model identifies a query as belonging to the extrinsic task space in Eq.~\ref{eq:actionspace}, RemoteAgent does not attempt to generate dense spatial outputs directly with the central MLLM. Instead, it reformulates extrinsic inference as an executable tool invocation over an external expert library $\mathcal{E}$, as shown in Fig.~\ref{overview}. This design is motivated by the fact that dense Earth Observation tasks, such as semantic segmentation, referring expression segmentation, and change detection, demand precision-critical spatial outputs that are inherently mismatched with autoregressive text generation.


\begin{figure}[b]
  \centering
  \includegraphics[width=\linewidth]{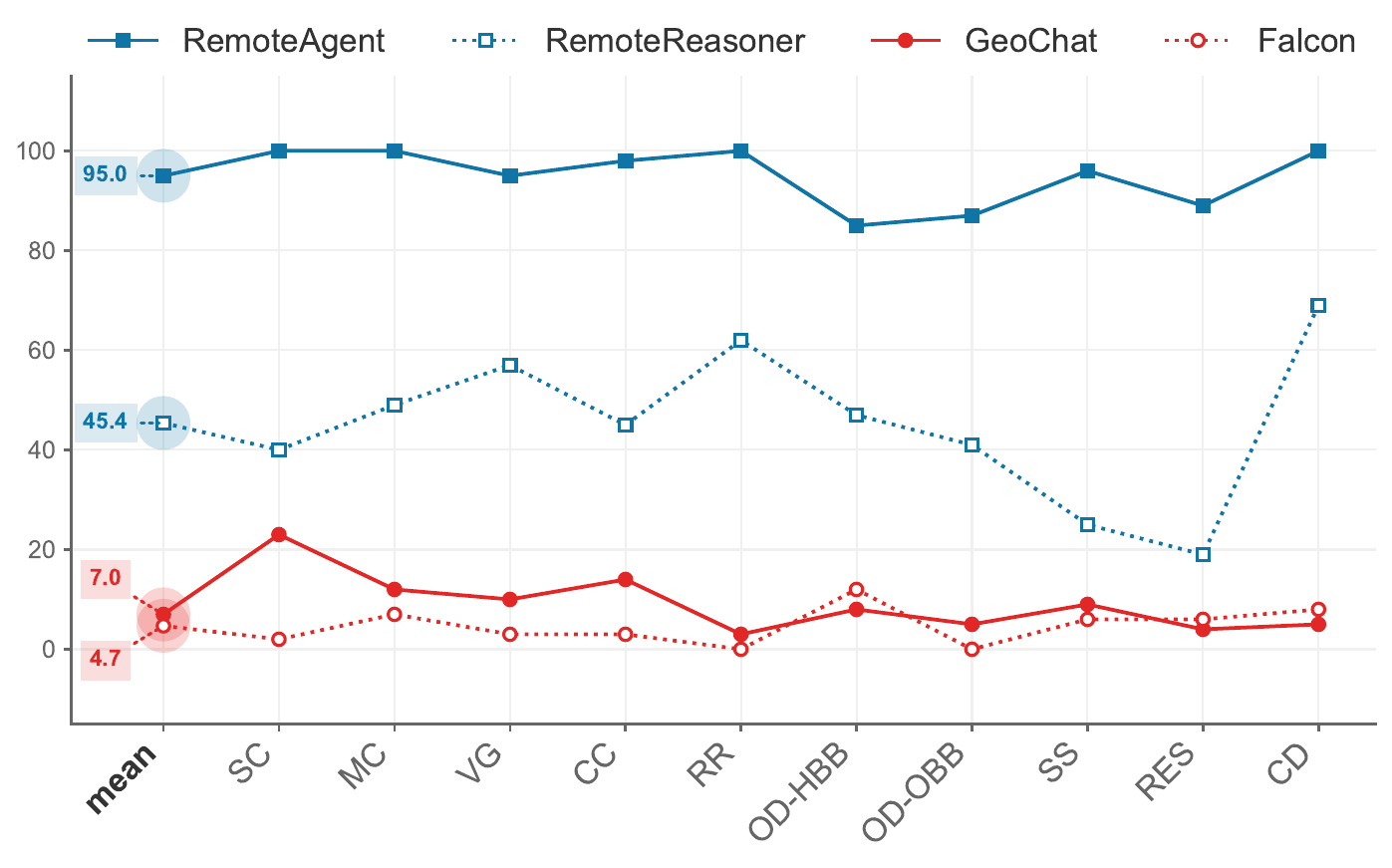}
  \caption{Intent recognition performance across diverse EO tasks on our VagueEO. RemoteAgent eclipses all baselines.}
  \label{intent}
\end{figure}

Formally, for an input pair $(I,Q)\in\mathcal{T}_{\mathrm{ex}}$, the policy model $\pi_{\theta}$ predicts both the target expert $e_k\in\mathcal{E}$ and its task-specific parameterization $p$:
\begin{equation}
(e_k,p)\sim \pi_\theta(\cdot \mid I,Q), \qquad \text{if } (I,Q)\in\mathcal{T}_{\mathrm{ex}}.
\end{equation}
The predicted pair $(e_k,p)$ is then instantiated as a structured tool call $T_{\mathrm{call}}(e_k,p)$, which serves as the explicit action emitted by the agent for extrinsic execution. In this way, the policy is responsible for high-level intent grounding and tool selection, rather than directly producing bounding boxes or masks token by token.

The generated instruction is dispatched through the Model Context Protocol (MCP), which provides a standardized interface between the central policy and heterogeneous specialized EO expert modules. After execution, the selected specialist returns the corresponding dense prediction $Y_{\mathrm{dense}}=e_k(p; I)$, where $Y_{\mathrm{dense}}$ may denote detection boxes or segmentation masks, depending on the invoked tool. This mechanism clearly decouples semantic reasoning from precision-sensitive spatial execution. The MLLM remains the cognitive core for interpreting vague human intent, while dense prediction is delegated only when the task exceeds its native output granularity. Consequently, RemoteAgent preserves the flexibility of the central model while achieving specialist-level execution on dense tasks.

%% file: sec/5_Experiments.tex
\section{Experiments}
\label{sec:experiments}

To rigorously validate our RemoteAgent, we evaluate its intent recognition capabilities on the VagueEO dataset while assessing its actual execution proficiency across established Earth Observation benchmarks. This section highlights a representative subset of tasks, specifically focusing on intent recognition, intrinsic sparse localization, and extrinsic dense spatial predictions. More experiments are deferred to the supplementary material.

\subsection{Experimental Setup}
We implement our reinforcement fine-tuning using the ms-swift~\cite{zhao2024swift} framework and DeepSpeed ZeRO-2~\cite{rasley2020deepspeed}. Initializing with Qwen2.5-VL-7B-Instruct~\cite{bai2025qwen2}, we apply LoRA ($r=32, \alpha=64$) across all linear layers. For the GRPO algorithm, we sample $G=4$ generations per query with a temperature of $0.95$. The model is trained for 24 epochs using a constant learning rate of $1\times 10^{-6}$ in bfloat16 precision, utilizing an effective batch size of 32 across 2 NVIDIA H100 GPUs. All tools are utilized with their official open-source implementations. In the MCP-based execution pipeline, all experts are encapsulated as MCP-compliant services and, together with the central MLLM, are deployed in a shared local environment with 8 NVIDIA 4090 GPUs.

\begin{table}[t]
\centering
\caption{Comparison of scene classification results. }
\renewcommand\arraystretch{1.07}
\resizebox{0.48\textwidth}{!}{
\begin{tabular}{lc|cc}
\toprule
\multirow{2}{*}{Methods}                   &\multirow{2}{*}{Publication} & AID~\cite{xia2017aid}   & WHU-RS19~\cite{balestra2025whu} \\
\cline{3-4}
                         &                              & $Acc$   & $Acc$    \\
\hline
InternVL3.5~\cite{wang2025internvl3}       &arXiv'25 & 73.80 & 91.50    \\
Qwen2.5-VL~\cite{bai2025qwen2}            &arXiv'25 & 63.07 & 76.60    \\
Phi3.5-Vision~\cite{abdin2024phi}             &arXiv'24& 56.57 & 68.90    \\
GeoChat~\cite{kuckreja2024geochat}                 &CVPR'24& 73.17 & 84.80    \\
EarthDial~\cite{soni2025earthdial}                &CVPR'25&87.57& \textbf{95.80}    \\
GeoMag~\cite{ma2025geomag}                   &MM'25& 83.03 & 77.62    \\
VHM~\cite{pang2025vhm}              &AAAI'25    & \uline{91.70} & \textbf{95.80}    \\
LHRS-Bot~\cite{muhtar2024lhrs}      & ECCV'24        & 91.26 & \uline{93.17}    \\
FUSE-RSVLM~\cite{dang2025fuse}               &arXiv'25& \textbf{94.37} & 93.10    \\
\hline \rowcolor{blue!8}
RemoteAgent               &- &  91.34	     &   90.23       \\
\bottomrule
\end{tabular}
}
\label{scenecls}
\end{table}


\begin{table}[t]
\centering
\caption{Comparison of visual grounding results. }
\resizebox{0.48\textwidth}{!}{
\begin{tabular}{lc|cc}
\toprule
\multirow{2}{*}{Methods} & \multirow{2}{*}{Publication} & \multicolumn{2}{c}{DIOR-RSVG~\cite{zhan2023rsvg}} \\
\cline{3-4}
                         &                              & $Acc@0.5$         & $IoU$        \\
\hline
SkyEyeGPT~\cite{zhan2025skyeyegpt}                & NIPS'22                       & 70.5           & -           \\
GeoChat~\cite{kuckreja2024geochat}                  & CVPR'24                      & 31.4            & 14.7        \\
SkySenseGPT~\cite{luo2024skysensegpt}              & arXiv'24                     & 60.8            & 35.5        \\
LHRS-Bot~\cite{muhtar2024lhrs}                 & ECCV'24                       & \textbf{73.5}           & -           \\
Falcon~\cite{yao2025falcon}                   & arXiv'25                      & 56.9          & -           \\
SkyMoE~\cite{liu2025skymoe}                   & arXiv'25                     & 68.6            & \textbf{48.6}        \\
VHM~\cite{pang2025vhm}                      & AAAI'25                      & 55.9            & 42.0        \\
EarthDial~\cite{soni2025earthdial}                & CVPR'25                      & 46.1            & 34.3        \\
\hline \rowcolor{blue!8}
RemoteAgent              & -                            & \uline{68.9}    &  \uline{48.3}           \\
\bottomrule
\end{tabular}
}
\label{tab:visual_grounding}
\end{table}

\subsection{Intent Recognition Results}
To verify whether our system bridges the usability gap, we first evaluate its prerequisite: deciphering ambiguous instructions. As Fig.~\ref{intent} shows, RemoteAgent achieves an overwhelming $95.0\%$ mean accuracy, completely eclipsing the RL-based model RemoteReasoner. In contrast, SFT-based MLLMs like GeoChat and Falcon nearly fail ($<8\%$), revealing that supervised fine-tuning tends to overfit models to rigid prompts and severely degrades semantic flexibility. This failure is also largely attributed to the scarcity of long, conversational prompts in their fine-tuning data. The result directly validates our two core design motivations. First, training on the VagueEO dataset explicitly exposes the model to the linguistic ambiguity inherent in real-world user queries. Crucially, our RL-based alignment circumvents the catastrophic forgetting typically induced by standard SFT. Rather than forcefully overwriting the MLLM's pre-trained language priors with rigid task templates, the RL paradigm acts as a lightweight steering mechanism, preserving the model's intrinsic reasoning capabilities while teaching it to route complex intents.

\subsection{Intrinsic Evaluations}

\subsubsection{Scene Classification}
Scene classification tests holistic macroscopic comprehension, a capability our agent must resolve intrinsically without external tool invocation. As summarized in Tab.~\ref{scenecls}, RemoteAgent demonstrates formidable internal visual perception, achieving an accuracy of 91.34 on the AID benchmark. This decisively eclipses general-purpose models like Qwen2.5-VL by over 28 points and heavily outperforms early remote sensing baselines like GeoChat. While trailing the absolute state-of-the-art specialist FUSE-RSVLM by a narrow margin, our framework remains exceptionally competitive across both datasets. This result confirms our training strategy successfully preserves MLLM's native image-level understanding capability.




\subsubsection{Grounding \& Reasoning}
As detailed in Tables \ref{tab:visual_grounding} and \ref{tab:region_reasoning}, RemoteAgent demonstrates highly competitive performance on visual grounding and geospatial region reasoning, establishing a strong overall trend against existing multi-modal large language models (MLLMs). Specifically, on the DIOR-RSVG dataset, RemoteAgent achieves an IoU of $48.3$, significantly surpassing baselines like EarthDial and Falcon. Similarly, in the region reasoning task, it delivers an $Acc@0.5$ of $57.81\%$ on the test set, outperforming Qwen2.5-VL-7B by a substantial margin of $16.6\%$. It validates that our framework successfully retains precise grounding and reasoning capabilities.

\begin{table}[t]
\centering
\caption{Comparison of geospatial region reasoning results with various MLLMs on EarthReason~\cite{li2025segearth}.}
\renewcommand\arraystretch{1.1}
\resizebox{0.48\textwidth}{!}{
\begin{tabular}{l|cc|cc}
\toprule
\multirow{2}{*}{Methods} & \multicolumn{2}{c|}{Test} & \multicolumn{2}{c}{Val} \\
\cline{2-5}
                         & $Acc@0.5$     & $Acc@0.5$    & $gIoU$       & $gIoU$       \\
                         \hline
DeepSeek-VL2-tiny~\cite{wu2024deepseek}        & 12.08       & 12.67      & 17.51      & 18.62      \\
GeoChat~\cite{kuckreja2024geochat}                  & 10.10 & 8.89       & 12.57      & 11.44      \\
Qwen2.5-VL-7B~\cite{bai2025qwen2}            & 41.21       & 45.82      & 38.77      & 41.80       \\
RemoteReasoner~\cite{yao2026remotereasoner}           & \textbf{66.51}       & \textbf{68.11}      & \textbf{67.04}      & \textbf{69.29}      \\
\hline \rowcolor{blue!8}
RemoteAgent              &   \uline{57.81}			 & \uline{54.22}           &  \uline{55.60}          & \uline{52.22}          \\
\bottomrule
\end{tabular}
}
\label{tab:region_reasoning}
\end{table}

\begin{table}[t]
\centering
\caption{Comparison of object counting results with various MLLMs on two datasets.}
\renewcommand\arraystretch{1.08}
\resizebox{0.48\textwidth}{!}{
\begin{tabular}{lc|cc}
\toprule
\multirow{2}{*}{Methods} & \multirow{2}{*}{Publication} & HRRSD~\cite{zhang2019hierarchical} & DOTAv2~\cite{xia2018dota} \\
\cline{3-4}
                         &                              & $Acc$   & $Acc$    \\
\hline
GeoChat~\cite{kuckreja2024geochat}                  & CVPR'24                      & 57.6  & 16.9   \\
VHM~\cite{pang2025vhm}                      & AAAI'25                      & 46.7  & 18.0   \\
RSUniVLM~\cite{liu2024rsunivlm}                & arXiv'24                     & 54.2  & 19.0   \\
LLaVA-1.5~\cite{liu2023visual}                & NIPS'24                      & -     & 22.1   \\
LHRS-Bot~\cite{muhtar2024lhrs}                 & ECCV'24                      & -     & 24.4   \\
EarthDial~\cite{soni2025earthdial}                & CVPR'25                      & \textbf{61.5}  & 20.9   \\
SkyMoE~\cite{liu2025skymoe}                   & arXiv'25                     & 57.8  & \uline{26.4}   \\
\hline \rowcolor{blue!8}
RemoteAgent              & -                            &  \uline{58.0}     &  \textbf{27.8}      \\
\bottomrule
\end{tabular}
}
\label{counting}
\end{table}

\subsubsection{Object Counting}
We also conducted the Object Counting task on two datasets. As shown in Tab.~\ref{counting}, the object counting evaluation further highlights the effectiveness of our RL-aligned model. RemoteAgent achieves SOTA performance on the DOTAv2 dataset, surpassing recent approaches such as SkyMoE and LHRS-Bot. On the HRRSD benchmark, it remains highly competitive, outperforming baselines including GeoChat and RSUniVLM. A small gap is observed compared to EarthDial on HRRSD.

\subsection{Extrinsic Evaluations}

\subsubsection{Object Detection}
Given the inherently dense distribution of remote sensing targets, object detection constitutes a dense prediction task that necessitates specialized external tools. We conduct a comparison of different models on both general detection and oriented detection in Tab.~\ref{det}. By routing these complex queries to dedicated detection tools, RemoteAgent drastically eclipses existing MLLMs, crushing Falcon by over 21 points on the DIOR benchmark and completely annihilating Florence-2-L. Furthermore, our framework rivals highly specialized detectors, trailing the state-of-the-art SkySense by less than one point across both DIOR and DIOR-R datasets. We attribute this marginal deficit entirely to a minute fraction of highly ambiguous queries misrouting during the initial intent recognition stage.

\subsubsection{Semantic Segmentation}
Semantic segmentation demands exhaustive pixel-level classification, a dense prediction format that overloads the text-generation bottleneck of standard MLLMs. To circumvent this, RemoteAgent intelligently delegates these types of queries to external segmentation experts. On the Potsdam benchmark, our framework achieves an outstanding 93.54 mF1, trailing only the absolute state-of-the-art SkySense while outperforming recent architectures like RS-vHeat. On the iSAID dataset, RemoteAgent yields a competitive 67.01 mIoU, maintaining a high level of performance consistent with its tool's native capabilities.

\begin{table}[t]
\centering
\caption{Comparison of object detection results with various specialized models and MLLMs.}
\resizebox{0.48\textwidth}{!}{
\begin{tabular}{lc|cc}
\toprule
\multirow{2}{*}{Methods} & \multirow{2}{*}{Publication} & DIOR~\cite{li2020object}  & DIOR-R~\cite{cheng2022anchor} \\
\cline{3-4}
                         &                              & $AP50$  & $AP50$              \\
\hline \rowcolor{gray!8}
\textit{Specialized Models}                    &                              &       &        \\
\hline
GFM~\cite{vatsavai2024geospatial}                     & ICCV'23                      & 72.84 & 67.67             \\
Scale-MAE~\cite{reed2023scale}                & ICCV'23                      & 73.81 & 66.47             \\
SkySense~\cite{guo2024skysense}                 & CVPR'24                      & \textbf{78.73} & \textbf{74.27}             \\
\hline \rowcolor{gray!8}
\textit{MLLMs}                    &                              &       &        \\
\hline
Florence-2-L~\cite{xiao2024florence}             & CVPR'24                      & 26.98 & -                 \\
Falcon~\cite{yao2025falcon}                   & arXiv'25                     & 56.65 & -                 \\
\hline \rowcolor{blue!8}
RemoteAgent              &          -                    & \uline{77.80} &  \uline{73.80}                \\
\bottomrule
\end{tabular}
}
\label{det}
\end{table}

\begin{table}[t]
\centering
\caption{Comparison of semantic segmentation results with various specialized models.}
\resizebox{0.48\textwidth}{!}{
\begin{tabular}{lc|cc}
\toprule
\multirow{2}{*}{Methods} & \multirow{2}{*}{Publication} & iSAID~\cite{waqas2019isaid} & Potsdam~\cite{sherrah2016fully} \\
\cline{3-4}
                         &                              & $mIoU$  & $mF1$     \\
\hline
Scale-MAE~\cite{reed2023scale}                & ICCV’23                      & 65.77 & 91.54   \\
MA3E~\cite{li2024masked}                   & ECCV’24                      & 64.06 & 91.50   \\
SkySense~\cite{guo2024skysense}                 & CVPR'24                      & \textbf{70.91} & \textbf{93.99}   \\
RS-vHeat~\cite{hu2025rs}                 & ICCV'25                      & \uline{68.72} & 92.82   \\
RemoteSAM~\cite{yao2025remotesam}                & MM'25                        & 64.72 & 91.80   \\
\hline \rowcolor{blue!8}
RemoteAgent              & -                            & 67.01      & \uline{93.54}        \\
\hline
\end{tabular}
}
\label{seg}
\end{table}

\subsubsection{Referring Expression Segmentation}
Referring expression segmentation also demands rigorous pixel-level precision. Therefore, our RemoteAgent dynamically delegates these dense spatial queries to a dedicated expert tool, RemoteSAM via MCP. The evaluation results in Tab.~\ref{RES} demonstrate the overwhelming advantage of this routing strategy. Our framework achieves state-of-the-art performance on the RRSIS-D benchmark, recording a peak $mIoU$ of $71.08$ and an $Acc@0.5$ of $83.64$. This significantly eclipses both specialized segmentation architectures, outperforming RS2-SAM2 by $4.36$ $mIoU$, and MLLM-based models like SegEarth-R2 ($+3.18$ $mIoU$). This performance confirms that intelligently orchestrating specialized tools for dense tasks is a far superior paradigm compared to forcing a single MLLM to generate dense outputs.

\begin{table}[t]
\centering
\caption{Comparison of referring expression segmentation results with various specialized models and MLLMs.}
\resizebox{0.48\textwidth}{!}{
\begin{tabular}{lc|ccc}
\toprule
\multirow{2}{*}{Methods} & \multirow{2}{*}{Publication} & \multicolumn{3}{c}{RRSIS-D~\cite{liu2024rotated}} \\
\cline{3-5}
                         &                              & $Acc@0.5$   & $oIoU$   & $mIoU$   \\
\hline \rowcolor{gray!8}
\textit{Specialized Models}                    &                      &        &       &        \\
\hline
LAVT~\cite{yang2022lavt}                     & CVPR'22                      & 69.52     & 77.19  & 61.04  \\

LGCE~\cite{yuan2024rrsis}                     & TGRS'24                      & 67.65     & 76.34  & 59.37  \\
RMSIN~\cite{liu2024rotated}                    & CVPR'24                      & 74.26     & 77.79  & 64.20  \\
CroBIM~\cite{dong2024cross}                   & TGRS'24                       & 74.58     & 75.99  & 64.46  \\
LGCE~\cite{yuan2024rrsis}                      & TGRS'24                       & 67.65     & 76.34  & 59.37  \\
RS2-SAM2~\cite{rong2025rs2}                 & AAAI'26                      & 77.56     & 78.99  & 66.72  \\
\hline \rowcolor{gray!8}
\textit{MLLMs}                   &                              &           &        &        \\
\hline 
LISA~\cite{lai2024lisa}                     & CVPR'24                      & 24.51     & -      & 26.78  \\
PixelLM~\cite{ren2024pixellm}                  & CVPR'24                      & 28.81     & -      & 31.65  \\
NEXT-Chat~\cite{zhang2023next}                & arXiv'23                     & 26.37     & -      & 24.98  \\
GeoGround~\cite{zhou2024geoground}                & arXiv'24                      & 67.50     & -      & 60.50  \\
SegEarth-R1~\cite{li2025segearth}              & arXiv'25                      & 76.96     & 78.01  & 66.40  \\
SegEarth-R2~\cite{xin2025segearth}              & CVPR'26                       & -         & -      & \uline{67.90}  \\
GeoPixel~\cite{shabbir2025geopixel}                 & ICML'25                      & -         & -      & 67.30  \\
Text4Seg++~\cite{lan2025text4seg++}              & ICLR'25                      & -         & -      & 62.80  \\
GeoMag~\cite{ma2025geomag}                  & MM'25                        & \uline{81.30}     & \textbf{82.67}  & 65.71  \\
\hline \rowcolor{blue!8}
RemoteAgent              & -                            & \textbf{83.64}     & \uline{79.50}  & \textbf{71.08}  \\
\bottomrule
\end{tabular}
}
\label{RES}
\end{table}

\begin{table}[t]
\centering
\caption{Comparison of building damage assessment results with various specialized models.}
\resizebox{0.48\textwidth}{!}{
\begin{tabular}{lc|ccc}
\toprule
\multirow{2}{*}{Methods} & \multirow{2}{*}{Publication} & \multicolumn{3}{c}{xBD} \\
\cline{3-5}
                         &                              & $F1_{loc}$                 & $F1_{cls}$                 & $F1_{overall}$                             \\
\hline                         
ChangeOS~\cite{zheng2021building}                 & RSE'21                       & \uline{85.69}            & 71.14                    & 75.5                         \\
DamFormer~\cite{chen2022dual}                & IGARSS'22                    & \textbf{86.86}           & 72.81                    & 77.02                        \\
PCDASNet~\cite{wang2024pcdasnet}                 & TGRS'24                      & 85.48                    & \textbf{73.83}           & \textbf{77.33}               \\
\hline \rowcolor{blue!8}
RemoteAgent              &            -                  & 80.12                    & \uline{73.03}              & \uline{77.16}              \\
\bottomrule
\end{tabular}
}
\label{BDA}
\end{table}


\subsubsection{Building Damage Assessment}
Building damage assessment inherently demands precise, bi-temporal pixel-level alignment to detect fine-grained structural change (a type of change detection). 
To better execute this task, RemoteAgent strategically routes such disaster evaluation queries to a dedicated change detection expert tool via the Model Context Protocol. The evaluation on the xBD benchmark in Tab.~\ref{BDA} highlights the efficacy of this delegation. Our framework achieves a highly competitive $F1_{overall}$ of $77.16$ and $F1_{cls}$ of $73.03$, surpassing established architectures like DamFormer and ChangeOS, albeit with a noticeable performance gap in the pure localization metric $F1_{loc}$ relative to PCDASNet. 
Ultimately, these results demonstrate that our agentic routing paradigm successfully extends the system's capabilities to complex, multi-temporal analytical tasks.

\subsection{Further Analysis}


\subsubsection{Ablation on Training Strategy}

To validate our training paradigm, we evaluate different training strategies in Tab.~\ref{ablation}. While SFT improves visual grounding, it triggers catastrophic forgetting in tool orchestration capability, plunging segmentation performance by 18.94\% mIoU compared to zero-shot baselines. Conversely, our reinforcement learning approach completely prevents this degradation, restoring segmentation to 71.64\% mIoU. Furthermore, RL delivers massive cognitive gains, outperforming SFT by 14.7 points in grounding and an overwhelming 28\% in intent accuracy. This definitely proves RL enhances multi-granularity execution without destroying intrinsic routing flexibility.


\begin{table}[t]
\caption{Ablation on different training strategies.} 
\resizebox{0.48\textwidth}{!}{
\begin{tabular}{l|cccc}
\toprule
Method & VG ($Acc@0.5$)     & RES ($mIoU$)  &  Intent ($Acc$) &Time ($s$)  \\
\hline
Zero-shot               & 43.6    & 71.13&  49          &0.84\\
SFT                     & 54.2    & 52.19&  67          &0.71\\
\hline \rowcolor{cyan!8}
\textbf{RL}                    & \textbf{68.9}    & \textbf{71.64} &  \textbf{95}     &0.83 \\
\bottomrule
\end{tabular}
}
\label{ablation}
\end{table}

\begin{table}[t]
\centering
\caption{Comparison of inference time efficiency.}
\resizebox{0.48\textwidth}{!}{
\begin{tabular}{l|ccc}
\toprule
Method                     & LLM (s) & Tool (s)& Total (s) \\
\hline
Earth-Agent (GPT)~\cite{feng2025earth}           & 158      & 42        & 200        \\
Earth-Agent (DeepSeek-V3.1)~\cite{feng2025earth}  & 51       & 28        & 79         \\
Earth-Agent (KimiK2)~\cite{feng2025earth}         & 105      & 27        & 132        \\
\hline  \rowcolor{cyan!8}
Ours                       & \textbf{0.84}      & \textbf{0.34}       & \textbf{1.18}          \\
\bottomrule
\end{tabular}
}
\label{time}
\end{table}

\begin{figure*}[t]
  \centering
  \includegraphics[width=\linewidth]{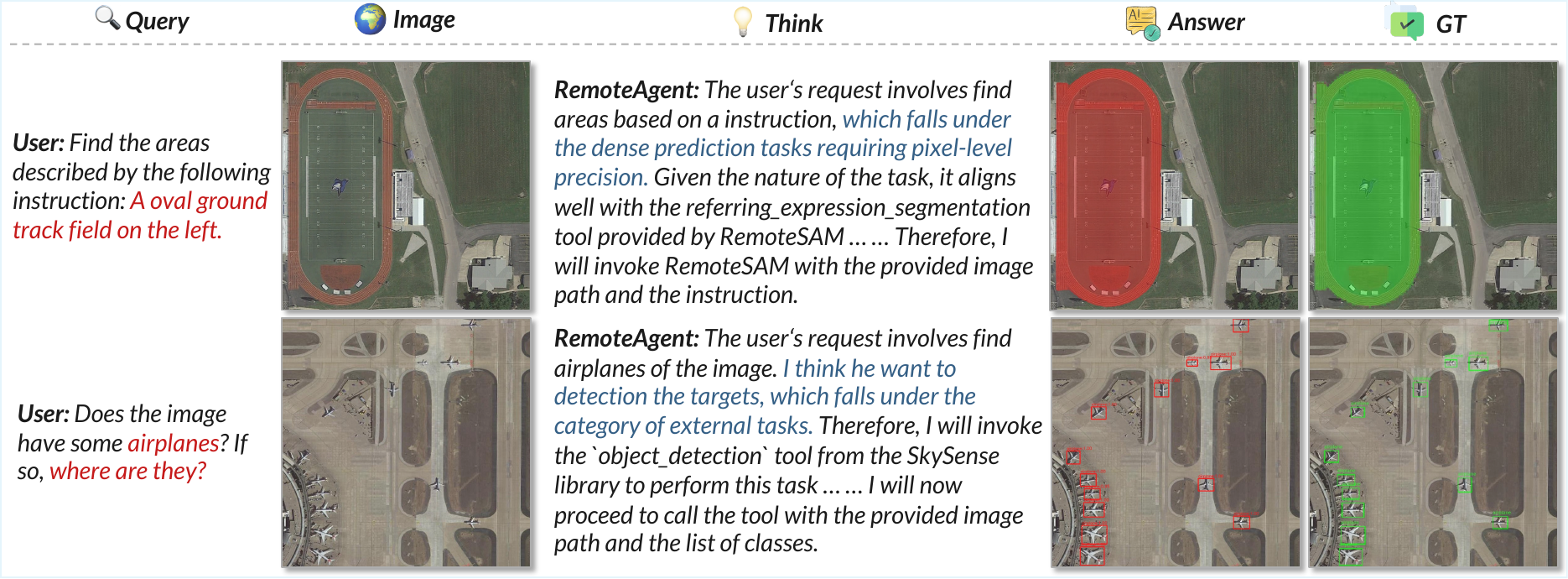}
  \caption{Qualitative results of RemoteAgent. The agent accurately interprets free-form queries and dynamically routes them to specialized tools, seamlessly bridging vague intents with precision-critical execution.}
  \label{vis}
\end{figure*}

\subsubsection{Time Efficiency}
Real-world deployments demand real-time responsiveness, a metric where current agentic frameworks severely falter. As illustrated in Tab.~\ref{time}, existing agentic systems like Earth-Agent rely on multi-step ReAct~\cite{yao2022react} reasoning loops, resulting in agonizing inference delays ranging from 79 seconds with DeepSeek-V3.1 to a staggering 200 seconds with GPT. Conversely, RemoteAgent achieves a lightning-fast total execution of just 1.18 seconds. By leveraging our robust intent recognition for direct, single-step tool invocation, we completely bypass redundant reasoning cycles, delivering an unprecedented 100x speedup without sacrificing execution precision. 

\subsubsection{Case Studies}
Real-world usability hinges on translating ambiguous queries into actionable execution workflows. In Fig.~\ref{vis}, we present qualitative cases demonstrating the dynamic routing capabilities of our framework. When tasked with locating an "oval ground track field" or identifying "airplanes", the agent's internal reasoning exhibits remarkable clarity. It autonomously recognizes the necessity for dense spatial outputs, accurately delegating the respective queries to RemoteSAM for pixel-wise referring segmentation and SkySense for object detection. It definitely confirms that RemoteAgent successfully maps free-form human intents to precise expert tools without manual intervention.

%% file: sec/2_related_work.tex
\section{Related Work}
\subsection{Remote Sensing MLLMs}

The integration of Multi-modal Large Language Models (MLLMs) into remote sensing has significantly advanced Earth observation. Initial efforts primarily adapted general-domain VLMs via large-scale instruction tuning for fundamental tasks such as image captioning and visual question answering~\cite{hu2025rsgpt, kuckreja2024geochat, zhang2024earthgpt, zhan2025skyeyegpt, yao2025falcon,li2024language}, which later evolved to support multi-granularity localization, temporal analysis, and fine-grained attribute comprehension~\cite{zhang2024earthmarker, wang2024ringmogpt, liu2024rsunivlm, irvin2024teochat, jiang2025eaglevisionobjectlevelattributemultimodal}. However, traditional MLLMs often struggle with complex spatial logic due to their direct end-to-end mapping paradigm. Consequently, a recent paradigm shift has emerged towards explicit geospatial reasoning driven by reinforcement learning (RL). Models such as Geo-R1~\cite{zhang2025geo}, RemoteReasoner~\cite{yao2026remotereasoner}, and RSThinker~\cite{liu2025towards} leverage RL to generate verifiable Chain-of-Thought (CoT) rationales prior to task execution. Pushing this boundary further, advanced frameworks now integrate task-aware rewards for pixel-level reasoning~\cite{li2025segearth, fiaz2025geovlm} and incentivize logical reasoning from scratch without predefined CoT supervision~\cite{wang2025geozero, li2026georeason}, aiming to resolve implicit queries and mitigate logical hallucinations in complex geospatial scenarios.
However, despite their strong semantic understanding, the inherently text-centric output format of existing MLLMs renders them ill-suited for dense, precision-critical spatial predictions in real-world remote sensing applications.

\subsection{Remote Sensing Agentic Systems}
Recent advancements have increasingly explored Large Language Models (LLMs) and Multi-modal Large Language Models (MLLMs) to automate complex remote sensing workflows. For instance, RS-Agent~\cite{xu2024rs} integrates a central controller with a dynamic toolkit and specialized knowledge spaces to autonomously orchestrate expert models, while GeoFlow~\cite{bhattaram2025geoflow} focuses on generating agentic workflows by providing detailed tool-calling objectives during runtime. Further expanding these capabilities, Earth-Agent~\cite{feng2025earth} unifies RGB and spectral data within an MCP-based ecosystem for cross-modal spatiotemporal reasoning, and OpenEarthAgent~\cite{shabbir2026openearthagent} aligns models with verified multi-step tool interactions through supervised fine-tuning. To manage intricate task dependencies, frameworks like EarthAgent~\cite{li2025designing} and CangLing-KnowFlow~\cite{chen2025cangling} introduce hierarchical task abstractions and expert-validated procedural knowledge bases to ensure logical completeness, supported by specialized evaluation benchmarks~\cite{shabbir2025thinkgeo}. 
Despite these strides, a critical limitation persists: these paradigms typically employ a rigid execution pipeline that treats the central model primarily as a dispatcher. By relying heavily on external tool chains even for rudimentary visual queries, they incur unnecessary computational overhead and latency. 


%% file: sec/6_Conclusion.tex
\section{Limitations \& Future Work}

Despite its success in bridging the usability gap in Earth Observation, RemoteAgent still faces a lot of limitations.
First, the scale of the VagueEO dataset is relatively limited and cannot exhaustively cover the distribution of real-world vague queries. Second, the external tool orchestration relies on a manually constructed, static library, lacking a dynamic mechanism to autonomously discover and integrate emerging specialist models. Finally, RemoteAgent is susceptible to compounding errors from external tools without a built-in self-correction or rollback mechanism. Future work will focus on scaling instruction data and developing open-ended, dynamic tool integration to further enhance robustness.

\section{Conclusion}
In this work, we directly tackle the persistent usability gap in Earth Observation, introducing VagueEO to ground ambiguous, non-expert queries. We also propose RemoteAgent, an agentic framework that leverages reinforcement fine-tuning to resolve intrinsic macroscopic tasks while intelligently routing dense predictions to specialized tools via MCP. Extensive evaluations confirm its exceptional data efficiency and expert-level precision, establishing a robust paradigm for highly accessible, human-centric EO systems.